\documentclass[letterpaper]{article} 
\usepackage[preprint]{aaai2027}  
\usepackage[hyphens]{url}  
\usepackage{graphicx} 
\usepackage{natbib}  
\usepackage{caption} 
\usepackage{todonotes}

\usepackage{algorithm}
\usepackage{algorithmic}
\usepackage{booktabs}
\usepackage{multirow}
\usepackage{listings}
\lstdefinestyle{prompt}{basicstyle=\footnotesize\ttfamily,breaklines=true,breakatwhitespace=true,columns=fullflexible,keepspaces=true,frame=single,framesep=4pt,xleftmargin=2pt,aboveskip=5pt,belowskip=3pt}

\title{Cybersecurity Detection Classification with Reasoning-enabled Language Models}

\author{
    Amol Khanna\corresponding, Manu Nandan, Cristian Viorel Popa, Joan Pujol-Roig, Diana Bolocan,\\
    Laura Vasilie, Alexandru Apostu, Chase Helwig, Mihaela Gaman, Michael Brautbar,\\
    Edward Raff, Chase Midler, Sven Krasser\corresponding
}
\affiliations{
    CrowdStrike\\
    \texttt{amol.khanna@crowdstrike.com; sven@crowdstrike.com}
}

\begin{document}

\maketitle

\begin{abstract}
A major issue in Security Operations Centers (SOCs) is alert fatigue, as the number of detections reported is more than staff can triage in a given day. Prior work prompts or fine-tunes large language models (LLMs) to emit a triage label directly, but does not train them to reason about whether a detection is a genuine threat. We train a chain-of-thought (CoT) reasoning-enabled triage classifier on real, human-labeled Windows endpoint detections by combining automated prompt optimization, self-training, and reinforcement learning with verifiable rewards. We find that CoT reasoning also degrades the label-token probabilities that automated triage relies on, so we separately train a calibrator that reads the full reasoning trace and estimates the probability that the verdict is correct. Our system reaches 82.6\% test accuracy and, at the high-confidence operating point that governs automated triage, improves benign recall by 43.0\% and malicious recall by 18.3\% over a direct-label LLM classifier. We further show that the trained calibrator is necessary---an untrained confidence judge collapses high-confidence recall to zero---and that a finetuned 30B model significantly outperforms frontier general-purpose models, motivating targeted training over scale.
\end{abstract}


\section{Introduction}

Modern corporate cybersecurity software produces an unrelenting stream of detections in an effort to surface both clearly malicious and potentially suspicious behavior. In large organizations, these detections are processed by Security Operations Centers (SOCs), which triage detections as malicious or benign. While analysts are essential in SOCs, human effort cannot scale with the ever-expanding domain of cybersecurity sensors and the growing technological footprint of companies~\cite{alahmadi2022}. Moreover, triage is costly: an analyst spends on the order of 1 hour reviewing each detection, so analyst labor---not detection volume alone---bounds how much of the stream a SOC can review. As a result, automating triage and making it more efficient are long-standing pursuits in the cybersecurity industry~\cite{hassan2019nodoze}.

Triage is a reasoning-driven task, but existing research prompts and post-trains large language models (LLMs) to directly produce a label after reading a detection~\cite{habibzadeh2025llmsoc,daniel2024labeling}. This approach takes advantage of pre-trained LLMs' wide range of knowledge and produces a measurable confidence for each classification, which is critical for effective automated triage; however, it does not utilize LLMs' inference-time reasoning abilities. We seek to bring a reasoning-centric approach to detection triage. \textbf{In this work, we train an LLM-based triage classifier which uses chain-of-thought (CoT)}~\cite{wei2022cot}. CoT explicitly elicits intermediate reasoning from an LLM before producing an answer, and consistently yields large performance gains across mathematical, logical, and symbolic tasks~\cite{suzgun2022bbh}. In addition to training this classiifer, we identify that CoT interferes with an LLM's ability to output classification confidence~\cite{tian2023calibration,xiong2023confidence}. This is critical for managing false positive and false negative risks, so we train a second model to estimate confidence. Results indicate that our system significantly outperforms both LLM-based classifiers which do not use CoT and base LLMs using CoT which have not been finetuned on cybersecurity detections. At the high-precision operating point that governs automated triage, our system improves benign recall by 43.0 and malicious recall by 18.3 percentage points over a direct-label classifier while reaching 82.6\% overall accuracy.

\noindent Concretely, we make the following contributions:
\begin{itemize}
    \item To our knowledge, the first detection-triage classifier explicitly trained to produce \emph{CoT reasoning} about whether a security detection is malicious or benign before emitting a label. 
    \item A four-stage training recipe---prompt optimization, self-training, reinforcement learning with verifiable rewards, and confidence calibration---in which each stage improves the calibrated high-confidence operating point.
    \item Experiments demonstrating that training the confidence calibrator is \emph{necessary}: an untrained judge collapses high-confidence recall to zero.
    \item Evidence that a domain-specialized 30B model outperforms general-purpose models on this task, motivating targeted training over scale.
\end{itemize}

\section{Related Work}

\subsection{Cybersecurity Triage}

Cybersecurity detection triage has a long lineage. Early systems matched host and network activity against hand-written signatures and rules~\cite{roesch1999snort}, and later work added alarm clustering, learned classifiers, and anomaly detectors to cut false positives and prioritize alerts~\cite{julisch2003clustering,pietraszek2004adaptive}. Such data-driven detectors proved hard to operationalize, however: anomaly-based methods contend with a wide semantic gap and extreme class imbalance~\cite{sommer2010outside} and degrade under the temporal and spatial distribution shift typical of security data~\cite{pendlebury2019tesseract}. Deep sequence models~\cite{du2017deeplog,vanede2022deepcase} and provenance-graph methods~\cite{hassan2019nodoze,milajerdi2019holmes} later improved coverage and reduced alert fatigue, yet, like their predecessors, they emit an opaque verdict rather than the human-readable reasoning an analyst can audit. 

Large language models (LLMs) have also been applied across the SOC~\cite{habibzadeh2025llmsoc,zhang2024llmcyber}. The most direct approach prompts or fine-tunes an LLM to read a detection and emit a label, as when \citet{daniel2024labeling} classify intrusion-detection rules by MITRE ATT\&CK technique. Others use the LLM as an explanation layer over a separate detector, attaching natural-language rationales to an anomaly classifier~\cite{ali2023huntgpt} or applying it to log-based anomaly detection~\cite{qi2023loggpt}. A growing agentic line instead embeds LLMs in the analyst workflow---co-teaming with human operators~\cite{albanese2025coteaming}, filtering false positives from static-analysis alerts~\cite{xiong2026sifting}, and serving as benchmarks for incident-analysis tasks including triage~\cite{jajodia2026siabench}. In nearly all of these, however, the LLM either labels a detection directly or narrates a verdict reached elsewhere; it is not trained to reason about whether the detection is itself a true or false positive. 

The most mature LLM triage systems instead target IT and cloud operations: RCACopilot routes incidents and predicts root-cause categories with explanatory narratives~\cite{chen2023rcacopilot}, and PACE-LM augments such predictions with calibrated confidence estimates~\cite{zhang2023pacelm}. This work foreshadows two ideas central to our approach---reasoning over an incident and attaching a trustworthy confidence---but does not carry them into security-detection triage. We close this gap: to our knowledge, we are the first to show that explicitly training a model to reason about security detections yields significant gains in triage accuracy, paired with a dedicated calibrator that estimates the probability the verdict is correct. 

\subsection{Training LLMs to Produce CoT}

Many works have developed methods to train LLMs to reason through specific tasks. Our approach, presented in the next section, employs four stages: prompt optimization, self-training, reinforcement learning, and confidence calibration. We provide key background for each of these stages here.

\paragraph{Prompt Optimization} LLM performance is highly sensitive to prompt wording, yet manual prompt engineering is brittle, labor-intensive, and difficult to reproduce. Automated optimizers instead search the instruction space: APE selects among model-generated candidates~\cite{zhou2023ape}, OPRO iteratively conditions on previously scored prompts~\cite{yang2024opro}, and EvoPrompt evolves a population with LLM-guided mutation and crossover~\cite{guo2024evoprompt}. We adopt GEPA~\cite{agrawal2025gepa}, which mutates prompts using natural-language reflection over execution traces to make targeted edits and reports state-of-the-art performance across a range of tasks; it is distributed as part of the DSPy framework~\cite{khattab2023dspy}.

\paragraph{Self-training} Self-training methods improve a model's reasoning using its own generations rather than human-written chains of thought. The foundational method, STaR~\cite{star2022}, samples rationales from the model and retains those that reach the correct answer. For samples the model gets wrong, it \textit{rationalizes}, generating a reasoning trace conditioned on the ground-truth answer; the model is then fine-tuned on all retained rationales, and the process proceeds iteratively. A variety of methods extend STaR, generalizing it beyond curated question answering~\cite{zelikman2024quietstar}, training verifiers on the solutions it discards~\cite{hosseini2024vstar}, and reframing self-training as offline reinforcement learning or expectation--maximization over self-generated data~\cite{gulcehre2023rest,singh2024restem,yuan2023rft,phan2023trice}. Among these, AdaSTaR augments the STaR loop with an adaptive difficulty curriculum so that training effort concentrates on the cases a model has not yet mastered~\cite{koh2025adastar}.

\paragraph{Reinforcement Learning} Reinforcement learning was first applied to LLMs for alignment, optimizing a learned model of human preferences with PPO~\cite{ouyang2022instructgpt,schulman2017ppo}. Recent post-training instead targets reinforcement learning with verifiable rewards (RLVR), in which outputs are checked for correctness programmatically---for instance, in math, coding, and classification problems. DeepSeek introduced the GRPO algorithm for this setting~\cite{deepseekmath2024} and demonstrated that large-scale RLVR elicits sophisticated reasoning, even directly from a base model~\cite{deepseekr1}. Subsequent reports corroborated these gains~\cite{kimiteam2025k15,lambert2024tulu3}, and simpler estimators such as RLOO remain competitive~\cite{ahmadian2024rloo}, sparking broad interest in RLVR for verifiable problems.

\paragraph{Confidence Calibration} Confidence calibration turns a model's raw outputs into a probability trustworthy enough to act on. Classical post-hoc methods rescale a classifier's scores---e.g., Platt scaling~\cite{platt1999} and temperature scaling~\cite{guo2017calibration}---but LLMs complicate the picture because their token- and sequence-level probabilities are themselves poorly calibrated, especially after instruction tuning~\cite{kadavath2022know,tian2023calibration}. Three families of confidence signal have emerged in response: verbalized methods prompt the model to state its certainty directly~\cite{lin2022teaching,xiong2023confidence}; sampling-based methods estimate confidence from the agreement of multiple stochastic generations~\cite{kuhn2023semantic}; and trained probes attach a separate predictor of correctness~\cite{kadavath2022know,cobbe2021verifiers}.

\section{Methods}
\label{sec:methods}

\subsection{Task Formulation}

We frame detection triage as binary classification. Each sample consists of a \textit{detection}, which is a multi-field JSON record describing a cybersecurity sensor record, and a \textit{label} which is either \texttt{malicious} or \texttt{benign}. The LLM classifier generates a reasoning trace and then a label in the format \texttt{<think>...</think>\textit{label}}, and the label is parsed from that output. We chose to finetune Nemotron-3-Nano-30B~\cite{nvidia2025nemotron3nano} for the classifier model. 

A detection aggregates heterogeneous fields an analyst would consider: process lineage (the command lines, file names, and file paths of the trigger process, its parent, and its grandparent), local and global prevalence, MITRE ATT\&CK tactic and technique mappings, detection metadata from the sensor (severity and pattern disposition), and free-text analyst context from related detections if available. Fields can be missing from detections.

\subsection{Prompt Optimization}

We began training with prompt optimization to yield the strongest possible starting point before the more expensive next stages. We used the GEPA optimizer, and extended this with a custom metric function and a token-limited instruction proposer. Nemotron-3-Super-120B~\cite{nvidia2026nemotron3super} was used as GEPA's reflection LLM. GEPA recommends using the strongest-possible LLM for reflection, but our legal environment restricted us to open-weight models approved by corporate security. To compensate, we modified GEPA's hyperparameters to use significantly more resources than the ``heavy'' defaults in DSPy. Details about prompt optimization, including the prompts we used for each LLM and hyperparameter modifications, can be found in the appendix. 

\paragraph{Custom Metric Function} In preliminary experiments using only accuracy as a metric, GEPA achieved $\sim$87.5\% accuracy but converged on decision-tree style instructions containing hard numeric thresholds (e.g., ``\texttt{if prevalance $\geq$ 70, output benign.}'') While these instructions scored well, they produce CoTs which are unhelpful to analysts, encode brittle rules with exceptions that degrade confidence, and undermine downstream confidence estimation. To prevent this, we implement an LLM-as-a-judge approach using Nemotron-3-Super-120B to score candidate instructions on five dimensions: multi-field reasoning, threshold rigidity, reasoning encouragement, field coverage, and generalization potential. Each dimension was rated from 0.0 to 1.0, and the metric function multiplied the \textit{minimum} of these five scores with accuracy, encouraging the model to output reasonable and comprehensive CoTs. 

\paragraph{Token-limited Instructions} Unconstrained instruction optimization tended to produce increasingly verbose system prompts. To prevent this, we implemented a custom GEPA proposal function that enforces a hard ceiling of 1,024 tokens on all candidate instructions. 

\subsection{Self-training}

After prompt optimization, we employed self-training to improve the model's accuracy. We built on AdaSTaR~\cite{koh2025adastar}, iteratively sampling from a priority heap which prioritizes difficult samples, generating reasoning traces with the partially trained model, verifying the reasoning traces for format consistency and correctness, rationalizing incorrect samples, performing low-rank (LoRA)~\cite{hu2022lora} finetuning, and updating the priority heap. We largely used AdaSTaR's open-source codebase, with significant modifications only for the rationalization component. Further details on self-training, including rationalization, can be found in the appendix. 

\paragraph{Rationalization} STaR-based methods~\cite{star2022} use rationalization to produce correct CoTs for samples which the partially finetuned model gets incorrect. This converts difficult examples into training data. Rationalization is typically achieved by prompting an LLM to read a sample with its correct output, and explain why this output is correct. However, when we tried this, we found the LLM would often produce outputs similar to ``\texttt{the answer is provided, and thus the correct answer is malicious.}'' To fix this, we developed a detailed rationalization system prompt and used LLM-as-a-judge scoring to only retain rationalizations which read as genuine independent analysis. 

\subsection{Reinforcement Learning}

Our classification problem can be framed as a verifiable reward environment. As such, we used GRPO~\cite{deepseekmath2024}, the canonical RLVR optimization algorithm, for further training from the self-training checkpoint. RLVR improves upon self-training since self-training only imitates training traces, while RLVR enables the model to discover novel reasoning strategies~\cite{deepseekr1}. Our reward function was 1.0 if and only if (a) the response follows the \texttt{<think>...</think>\textit{label}} format and (b) the predicted label matches the ground truth; otherwise, the reward is 0.0. Further details can be found in the appendix. 

\subsection{Confidence Calibration}

After RLVR, the policy model produces CoTs which commit to a conclusion before emitting a label token. For example, an LLM response may read ``\texttt{<think>...based on this evidence, the detection is malicious. Thus output malicious.</think>malicious}.'' This is natural in a CoT, but makes the label token (\texttt{malicious}) near-certain under the model's next token distribution. Therefore, the output probability of the label token is \textit{not} a useful confidence signal---it is pathologically overconfident for both correct and incorrect predictions. 

To address this, we trained a separate calibrator model which judges whether the LLM's prediction is correct~\cite{kadavath2022know,cobbe2021verifiers}. We finetuned another Nemotron-3-Nano-30B with LoRA using supervised finetuning to read a detection, CoT, and label, and output a single-token verdict: \texttt{correct} or \texttt{incorrect}. We could then extract confidence by taking the softmax of logits of these tokens, yielding a score in $[0, 1]$ that reflects the calibrator's certainty. Further details can be found in the appendix.

\section{Experiments}

\subsection{Setup and Metrics}

Our dataset consists of human-labeled Windows endpoint sensor detections collected from a large SOC at a cybersecurity company. The training/validation/testing datasets have 388,336/59,162/42,686 detections and were collected over consecutive 8/2/1 week periods, respectively. An example of a detection is in Figure~\ref{fig:qualitative}. All training and inference compute was performed with 32 B200 GPUs. Nvidia Automodel, Nvidia Megatron Bridge, and Nvidia Nemo RL were used for finetuning. 

Working with SOC analysts, we defined three metrics of interest: 
\begin{enumerate}
    \item Benign recall at 98\% (high) precision: at this precision, analysts feel comfortable automatically closing detections. A high recall indicates that many truly benign samples will be closed. 
    \item Malicious recall at 99\% (high) precision: at this precision, analysts feel comfortable prioritizing detections. A high recall indicates that many truly malicious samples will be prioritized. 
    \item Accuracy: the CoT and label from a model with high accuracy can be shown to analysts to improve efficiency. 
\end{enumerate}
In addition to these metrics, we also calculate recall at 95\% (medium) and 90\% (low) precision for benign samples and 90\% (medium) and 80\% (low) precision for malicious samples. These thresholds were provided by analysts as indicators of model confidence. 

\subsection{Training}

\paragraph{Prompt Optimization} We ran prompt optimization with 100 candidate equivalents, a reflection minibatch size of 8, up to 15 merge invocations, and 16 parallel threads. We used this setting to compensate for a weaker reflection model since it produces significantly more metric evaluations than the default ``heavy'' setting in DSPy. Our hand-written prompt had 62.0\% accuracy on the validation set. Prompt optimization without an LLM-as-a-judge guardrail for reasoning yielded a prompt with 87.5\% accuracy; however, the prompt produced with the guardrail yielded 72.0\% accuracy. 

\begin{figure*}[t]
\centering
\includegraphics[width=\textwidth]{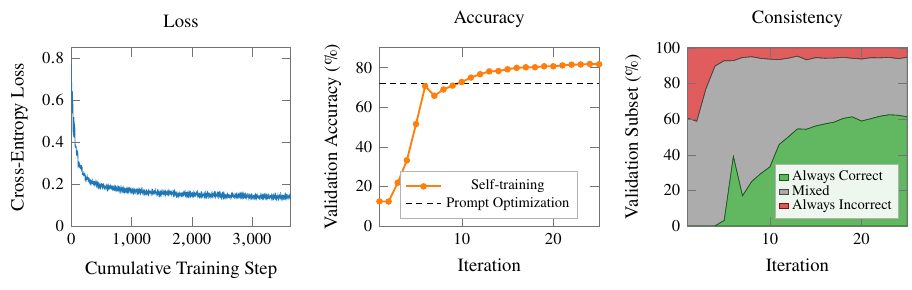}
\caption{Self-training dynamics across AdaSTaR iterations. \textbf{(a)} Cross-entropy training loss over cumulative steps drops sharply as the model adapts to the output format and stays stable (no spikes) across iterations. \textbf{(b)} Stochastic (temperature 1.0) validation accuracy against the GEPA-only prompt baseline (72.0\%, dashed). Accuracy is \emph{below} the baseline for the first several iterations---reflecting early training instability and overfitting before the model adapts---then surpasses it and peaks at 81.7\%. \textbf{(c)} Prediction consistency on a randomly selected 2{,}048-example validation subset: the always-incorrect fraction falls from 39.5\% to 5.4\% while the always-correct fraction grows.}
\label{fig:adastar}
\end{figure*}

\paragraph{Self-training} We self-trained the LLM for 25 iterations. The model was evaluated on the full validation set after every iteration, and a random subset of 2,048 samples was chosen to evaluate model consistency by generating 8 responses to each sample at temperature 1.0. Iteration 25 was chosen as the final checkpoint based on peak validation accuracy, but performance gains largely plateaued after iteration 18.

Figure~\ref{fig:adastar} summarizes the self-training run. Loss decreases rapidly in the first 100 steps, reflecting the model's initial adaptation to the expected output format. The absence of loss spikes confirms the optimizer-state accumulation across iterations provides stable training. The model's accuracy also improves by nearly 10 percentage points to 81.7\%, indicating that the training generalizes to the held-out validation set. Finally, the fraction of always incorrect examples when evaluating consistency decreases, indicating the model is capable of producing correct responses and is well-suited for GRPO. 

\begin{figure*}[t]
\centering
\includegraphics[width=\textwidth]{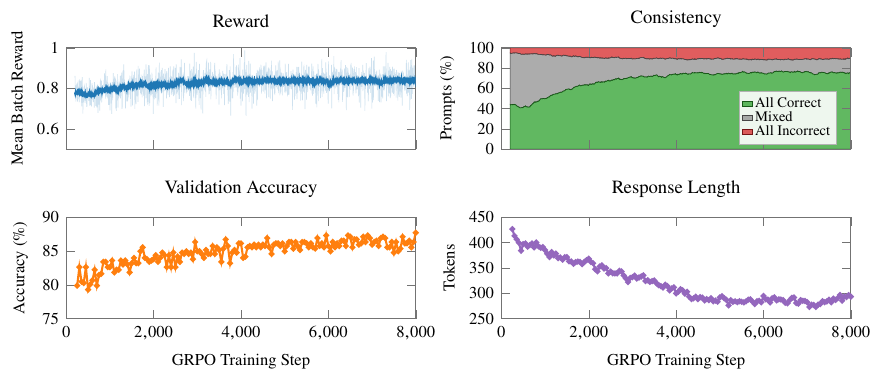}
\caption{GRPO reinforcement-learning dynamics over 8{,}000 training steps; all panels share the training step on the x-axis. \textbf{(a)} Mean batch reward (per-step in light, smoothed in bold) stays high and stable ($\sim$0.8) throughout. \textbf{(b)} Prompt consistency: the fraction of prompts whose sampled rollouts are \emph{all} correct grows from about half to roughly three-quarters as mixed-outcome prompts resolve, leaving a small residual set that remains consistently incorrect. \textbf{(c)} Validation accuracy rises to 87.7\%. \textbf{(d)} Mean validation response length falls from 426 to under 300 tokens by step $\sim$5{,}000 before leveling off, so the policy becomes both more accurate and more succinct.}
\label{fig:grpo}
\end{figure*}

\paragraph{Reinforcement Learning} We ran GRPO for 8000 steps, beginning with the Nemotron-3-Nano-30B base model merged with the self-training LoRA adapter. The model was evaluated on a randomly selected set of 512 validation samples after every 50 steps. Step 8000 was selected based on peak validation accuracy, but performance gains largely plateaued after step 5000. 

Figure~\ref{fig:grpo} summarizes the GRPO run. Reward remained consistent during training. Importantly, the fraction of samples which are consistently correct significantly rises during training. This is important because during deployment, a model which performs consistently is more trustworthy and easier to integrate. The model's validation accuracy also improves by 6 percentage points over the self-training checkpoint to 87.7\%. Interestingly, until step 5000, the model's mean validation response length decreases. This is also helpful during deployment, since it means the model became more accurate and succinct, which saves compute costs and human effort. 

\paragraph{Confidence Calibration} To train the calibrator, we generated 4 rollouts at temperature 1.0 for each sample in the training set. We then labeled samples as \texttt{correct} or \texttt{incorrect}, and trained the calibrator for 3 epochs. To choose which checkpoint to use, we identified the one which yielded the highest recall at 98\% precision for \texttt{benign} samples on the validation set. This recall was computed over samples which calibrator labels as \texttt{correct}, since the CoTs associated with these detections are consistent with calibration and can be surfaced to analysts. 

\subsection{Results}

\begin{figure*}[t]
\centering
\includegraphics[width=\textwidth]{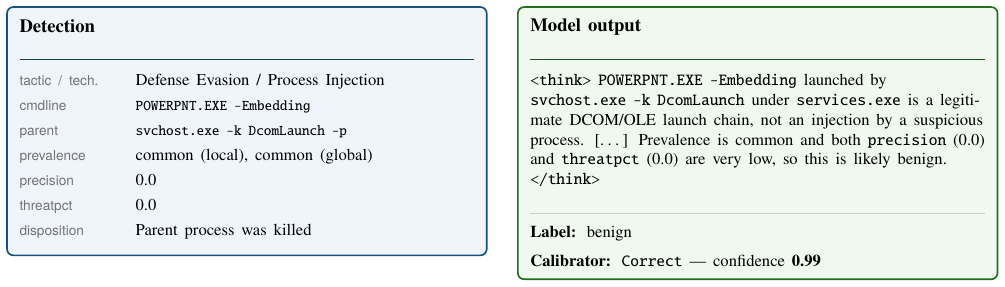}
\caption{A representative test-set detection (reasoning trace abridged). The classifier dismisses a severe-sounding \emph{process injection} alert as benign from the legitimate DCOM/OLE launch chain, and the calibrator endorses it with high confidence---so it can be auto-closed.}
\label{fig:qualitative}
\end{figure*}

\begin{table*}[t]
\centering
\small
\setlength{\tabcolsep}{4.5pt}
\renewcommand{\arraystretch}{0.95}
\begin{tabular}{lllrrrr}
\toprule
 & & & \multicolumn{2}{c}{Validation} & \multicolumn{2}{c}{Test} \\
\cmidrule(lr){4-5}\cmidrule(lr){6-7}
System & Label & Confidence & Precision\ (\%) & Recall (\%) & Precision\ (\%) & Recall (\%) \\
\midrule
\multirow{6}{*}{\shortstack[l]{Direct-label\\SFT}}
& \multirow{3}{*}{Benign}    & High   & 98.9 &  7.2 & 98.3 & 21.8 \\
&                            & Medium & 95.7 & 85.1 & 89.2 & 75.6 \\
&                            & Low    & 95.7 & 85.1 & 89.2 & 75.6 \\
\cmidrule(l){2-7}
& \multirow{3}{*}{Malicious} & High   & 99.0 & 23.8 & 98.3 & 34.7 \\
&                            & Medium & 96.1 & 36.1 & 98.1 & 45.7 \\
&                            & Low    & 85.0 & 87.6 & 89.7 & 88.6 \\
\midrule
\multirow{6}{*}{\shortstack[l]{\textbf{CoT+Calibration}\\\textbf{(ours)}}}
& \multirow{3}{*}{Benign}    & High   & 98.0 & 80.0 & 90.8 & \textbf{64.8} \\
&                            & Medium & 95.0 & 84.4 & 87.2 & 74.0 \\
&                            & Low    & 94.6 & 84.6 & 86.7 & 74.6 \\
\cmidrule(l){2-7}
& \multirow{3}{*}{Malicious} & High   & 99.0 & 46.5 & 98.9 & \textbf{53.0} \\
&                            & Medium & 92.6 & 79.1 & 92.5 & 82.2 \\
&                            & Low    & 92.6 & 79.1 & 92.5 & 82.2 \\
\bottomrule
\end{tabular}
\caption{Precision and recall for the direct-label SFT baseline and our CoT and calibration system at three precision-targeted confidence tiers.}
\label{tab:main-results}
\end{table*}

\paragraph{Overall Metrics} Table~\ref{tab:main-results} compares our full system against a \textit{direct-label} SFT baseline: a Nemotron-3-Nano-30B model trained to only output a label. The log-probabilities of this output can be used directly, since chain-of-thought does not interfere with probabilities~\cite{habibzadeh2025llmsoc,daniel2024labeling}. This represents the conventional automated triage approach and isolates the contribution of CoT reasoning and calibrator training.

The most important comparison is at the high confidence tier, which governs automated deployment. Our system achieves 64.8\% benign recall and 53.0\% malicious recall---gains of \textbf{+43.0} and \textbf{+18.3} percentage points over the direct-label baseline (benign 21.8\%, malicious 34.7\%). This improvement is attributable to the trained calibrator reading the GRPO policy's full reasoning trace, which provides a richer confidence signal than the detection alone: the calibrator can assess whether the stated reasoning supports the conclusion, endorsing a far larger fraction of correct predictions with high confidence. Figure~\ref{fig:qualitative} shows this end to end on a representative benign detection. However, it is important to note that benign high precision drifts from 98.0\% on validation to 90.8\% on test for our system, driven by the 4.4-point policy accuracy gap between the validation and test distributions; malicious high transfers nearly perfectly. By contrast, the direct-label baseline's benign high precision drifts by only 0.6 percentage points, demonstrating that the calibrator's precision advantage is partly offset by greater distribution sensitivity.

At medium and low tiers, both systems perform comparably, confirming that CoT training does not degrade coverage at lower confidence targets. However, in some instances, medium- and low-confidence precision/recall metrics are identical. This is because confidence distributions for predictions are bimodal---when the model is confident, its predictions tend to clear the high-confidence threshold, but when it is not confident, its predictions default to very low confidence.  

\begin{table}[t]
\centering
\scriptsize
\setlength{\tabcolsep}{3pt}
\renewcommand{\arraystretch}{0.95}
\begin{tabular}{lrrrrr}
\toprule
 & & \multicolumn{2}{c}{Benign High} & \multicolumn{2}{c}{Malicious High} \\
\cmidrule(lr){3-4}\cmidrule(lr){5-6}
System & Acc. & Precision & Recall & Precision & Recall \\
\midrule
Optimized Prompt           & 66.1          & 88.7          & 39.2          & 99.2          & 42.4 \\
+Self-training             & 76.0          & 90.8          & 52.2          & \textbf{99.7} & 44.1 \\
+RLVR (no self-training)              & 81.7          & \textbf{92.1} & 61.6          & 98.8          & 44.8 \\
\textbf{+Self-training+RLVR (ours)}   & \textbf{82.6} & 90.8          & \textbf{64.8} & 99.0          & \textbf{53.0} \\
\bottomrule
\end{tabular}
\caption{Ablating each finetuning stage (test set). All numbers are percentages. Each row's calibrator is trained independently, using the checkpoint that maximizes benign validation recall at 98\% precision. Benign- and malicious-high recall improve monotonically with each stage.}
\label{tab:policy-ladder}
\end{table}

\paragraph{Training Stage Ablation}

Table~\ref{tab:policy-ladder} ablates each training stage by comparing high-confidence performance after each step in the pipeline. The \emph{+self-training} row uses the above mentioned iteration~25 as the policy checkpoint---the run depicted in Figure~\ref{fig:adastar}---paired with a calibrator trained from scratch on that policy's rollouts. Each subsequent stage improves policy accuracy and benign recall monotonically.

Self-training raises benign recall by 13 percentage points with a modest malicious recall gain . The \emph{+RLVR (no self-training)} row applies GRPO directly to the GEPA-optimized base model, bypassing AdaSTaR entirely, using the same reward function as the full pipeline. This single-stage approach reaches 61.6\% benign recall---exceeding self-training alone---and lifts policy accuracy to 81.7\%, demonstrating the raw power of reinforcement learning from verifiable rewards on a well-prompted model. Training plots for this configuration are provided in the appendix.

Despite its strong benign recall, RLVR without self-training falls short on malicious recall. Combining both stages delivers the strongest performance on every axis, confirming that self-training and RL are complementary. Malicious precision remains stable near 99\% across all configurations, indicating that the calibrator reliably identifies high-confidence malicious predictions regardless of which policy stage is used.

\begin{table}[t]
\centering
\small
\setlength{\tabcolsep}{4pt}
\renewcommand{\arraystretch}{0.95}
\begin{tabular}{llrrr}
\toprule
 & & Val Benign & Test Benign & Test Benign \\
Policy & Calibrator & Recall & Recall & Precision \\
\midrule
Base      & Base      & 0.0           & 0.0           & --- \\
Base      & Finetuned & 67.9          & 39.2          & 88.7 \\
Finetuned & Base      & 0.0           & 0.0           & --- \\
Finetuned & Finetuned & \textbf{80.0} & \textbf{64.8} & \textbf{90.8} \\
\bottomrule
\end{tabular}
\caption{Ablating CoT training and calibrator training at the high-confidence benign precision target. All numbers are percentages. \textit{Base} is Nemotron-3-Nano-30B without finetuning.}
\label{tab:ablation-2x2}
\end{table}

\paragraph{CoT and Calibration Ablation}

Table~\ref{tab:ablation-2x2} isolates the contributions of the two key training decisions: CoT policy finetuning and calibrator training. \textit{Base} refers to Nemotron-3-Nano-30B with the GEPA-optimized system prompt but no parameter updates, and \textit{finetuned} to the trained counterpart (GRPO policy or trained calibrator, respectively). The calibrator for each \textit{finetuned} calibrator row was trained on that row's policy rollouts using the same pipeline as the full system.

The two zero-recall rows reveal a structural failure of an untrained confidence judge. The base model's confidence output clusters tightly near $\sigma(\pm1)\approx0.731$, regardless of whether the underlying policy is finetuned. Precision-targeted thresholds therefore land in a flat region of the confidence distribution with effectively no predictions above them, collapsing recall to zero in both cases. Calibrator training is therefore a \emph{necessary} component, not merely helpful: without it, neither policy quality nor reasoning accuracy is recoverable at high precision.

When paired with a finetuned calibrator, even the unfinetuned base policy achieves 39.2\% high-confidence benign recall on the test dataset. Policy finetuning then adds a further 25.6 percentage points, confirming that both training stages contribute meaningfully. The calibrator-training contribution dwarfs the policy-training contribution, underscoring that a well-trained confidence judge is the more critical factor, while a stronger CoT policy provides a complementary but substantial additional gain.

\begin{table}[t]
\centering
\scriptsize
\setlength{\tabcolsep}{3pt}
\renewcommand{\arraystretch}{0.95}
\begin{tabular}{lrrrrr}
\toprule
 & & \multicolumn{2}{c}{Benign} & \multicolumn{2}{c}{Malicious} \\
\cmidrule(lr){3-4}\cmidrule(lr){5-6}
Model & Acc. & Precision & Recall & Precision & Recall \\
\midrule
Nemotron-3-Nano-30B (base)   & 65.8 & 55.5 & 63.7 & 74.2 & 67.2 \\
Nemotron-3-Super-120B (base) & 62.1 & 51.0 & 80.6 & 80.1 & 50.3 \\
Claude Haiku 4.5             & 54.8 & 46.1 & 91.3 & 84.9 & 31.4 \\
Claude Sonnet 4.6            & 64.6 & 52.9 & 85.9 & 84.9 & 50.9 \\
Claude Opus 4.6              & 66.6 & 54.1 & 97.2 & 96.3 & 47.0 \\
\bottomrule
\end{tabular}
\caption{Five base models on a 1{,}000-sample test subset at $T{=}0$; All numbers are percentages. Precision/recall are computed from raw binary predictions with no confidence threshold, unlike the precision-targeted tiers in Table~\ref{tab:main-results}.}
\label{tab:baselines}
\end{table}

\paragraph{Comparison with Frontier Models}

Table~\ref{tab:baselines} compares the performance of five base models on the classification task. All models achieve broadly similar overall accuracy (55--67\%), and the top-ranked zero-shot model (Claude Opus) is only marginally ahead of the Nemotron-3-Nano-30B. \textit{Larger or more capable general-purpose models do not automatically transfer to this specialized domain}: Nemotron-3-Super underperforms Nano despite being a larger model, and Claude Opus---the strongest general-purpose model tested---roughly ties Nano on accuracy.

Precision and recall reveal a systematic bias in all Claude models: they predict benign with high recall (86--97\%) and rarely predict malicious, yielding very high malicious precision but correspondingly low malicious recall. Claude Opus, for example, predicts malicious on only 47\% of truly malicious detections, but is correct 96\% of the time when it does. Nemotron base is considerably more balanced---74\% malicious precision at 67\% recall.

Notably, benign precision is low across all models, indicating that benign predictions carry substantial uncertainty at zero shot: a significant fraction of detections labeled as benign are actually malicious. The trained CoT pipeline (Table~\ref{tab:policy-ladder}) achieves 82.6\% policy accuracy, a substantial gain over every zero-shot baseline, with the calibrated high-confidence tier further raising the actionable precision to over 90\%.

\section{Conclusion}

We presented a detection-triage system that is explicitly trained to reason about whether a security detection is a true or false positive, paired with a calibrator that turns the resulting reasoning trace into a trustworthy confidence. Trained with prompt optimization, self-training, and reinforcement learning on real Windows endpoint detections, the system improves high-confidence benign and malicious recall by 43.0 and 18.3 percentage points over a conventional direct-label classifier and reaches 82.6\% accuracy---well above every zero-shot baseline, including frontier models far larger than our 30B classifier.

This work addresses a severe bottleneck in the cybersecurity domain. The SOC we collaborated with processes nearly 1 million events in a single quarter, and scaling manual triage to meet this is impossible. It requires a massive team of expert analysts, which 1) is difficult to build and retain, 2) is prohibitively expensive, and 3) suffers from alert fatigue. By automating the high-confidence triage tier, our system mitigates these challenges and allows organizations to allocate their human capital to the most complex threats.

Two technical findings stand out. First, the training stages compound: prompt optimization, self-training, and reinforcement learning each lift the calibrated high-confidence operating point. Second, the calibrator is not an optional add-on but a prerequisite for high-precision automation---without it, high-confidence recall collapses to zero. We also observe that the benign high-precision operating point is sensitive to distribution shift, drifting from validation to test, which argues for continued monitoring and periodic recalibration in deployment.

Several directions remain. Our task is binary and limited to Windows endpoint detections; extending the approach to multi-class triage, additional sensor platforms, and tighter control of the validation-to-test precision gap are natural next steps toward broader operational use.


\bibliography{aaai2027}

@misc{star2022,
  title         = {{STaR}: Bootstrapping Reasoning With Reasoning},
  author        = {Zelikman, Eric and Wu, Yuhuai and Mu, Jesse and
                   Goodman, Noah D.},
  year          = {2022},
  eprint        = {2203.14465},
  archivePrefix = {arXiv},
}

@misc{deepseekmath2024,
  title         = {{DeepSeekMath}: Pushing the Limits of Mathematical Reasoning
                   in Open Language Models},
  author        = {Shao, Zhihong and Wang, Peiyi and Zhu, Qihao and Xu, Runxin
                   and Song, Junxiao and Bi, Xiao and Zhang, Haowei and
                   Zhang, Mingchuan and Li, Y. K. and Wu, Y. and Guo, Daya},
  year          = {2024},
  eprint        = {2402.03300},
  archivePrefix = {arXiv},
}

@inproceedings{alahmadi2022,
  title     = {99\% False Positives: A Qualitative Study of {SOC} Analysts'
               Perspectives on Security Alarms},
  author    = {Alahmadi, Bushra A. and Axon, Louise and Martinovic, Ivan},
  booktitle = {31st USENIX Security Symposium (USENIX Security 22)},
  pages     = {2783--2800},
  year      = {2022},
  publisher = {USENIX Association},
}

@inproceedings{hassan2019nodoze,
  title     = {{NoDoze}: Combatting Threat Alert Fatigue with Automated
               Provenance Triage},
  author    = {Hassan, Wajih Ul and Guo, Shengjian and Li, Ding and
               Chen, Zhengzhang and Jee, Kangkook and Li, Zhichun and
               Bates, Adam},
  booktitle = {Proceedings of the 26th Annual Network and Distributed System
               Security Symposium (NDSS)},
  year      = {2019},
}

@inproceedings{roesch1999snort,
  title     = {Snort: Lightweight Intrusion Detection for Networks},
  author    = {Roesch, Martin},
  booktitle = {Proceedings of the 13th USENIX Conference on System
               Administration (LISA '99)},
  pages     = {229--238},
  year      = {1999},
  publisher = {USENIX Association},
}

@inproceedings{sommer2010outside,
  title     = {Outside the Closed World: On Using Machine Learning for Network
               Intrusion Detection},
  author    = {Sommer, Robin and Paxson, Vern},
  booktitle = {2010 IEEE Symposium on Security and Privacy},
  pages     = {305--316},
  year      = {2010},
  publisher = {IEEE},
}

@inproceedings{pendlebury2019tesseract,
  title     = {{TESSERACT}: Eliminating Experimental Bias in Malware
               Classification across Space and Time},
  author    = {Pendlebury, Feargus and Pierazzi, Fabio and Jordaney, Roberto
               and Kinder, Johannes and Cavallaro, Lorenzo},
  booktitle = {28th USENIX Security Symposium (USENIX Security 19)},
  pages     = {729--746},
  year      = {2019},
  publisher = {USENIX Association},
}

@misc{habibzadeh2025llmsoc,
  title         = {Large Language Models for Security Operations Centers: A
                   Comprehensive Survey},
  author        = {Habibzadeh, Ali and Feyzi, Farid and
                   Ebrahimi Atani, Reza},
  year          = {2025},
  eprint        = {2509.10858},
  archivePrefix = {arXiv},
}

@misc{daniel2024labeling,
  title         = {Labeling {NIDS} Rules with {MITRE} {ATT\&CK} Techniques:
                   Machine Learning vs. Large Language Models},
  author        = {Daniel, Nir and Kaiser, Florian Klaus and Giladi, Shay and
                   Sharabi, Sapir and Moyal, Raz and Shpolyansky, Shalev and
                   Murillo, Andres and Elyashar, Aviad and Puzis, Rami},
  year          = {2024},
  eprint        = {2412.10978},
  archivePrefix = {arXiv},
}

@article{julisch2003clustering,
  title   = {Clustering Intrusion Detection Alarms to Support Root Cause
             Analysis},
  author  = {Julisch, Klaus},
  journal = {ACM Transactions on Information and System Security},
  volume  = {6},
  number  = {4},
  pages   = {443--471},
  year    = {2003},
}

@inproceedings{pietraszek2004adaptive,
  title     = {Using Adaptive Alert Classification to Reduce False Positives in
               Intrusion Detection},
  author    = {Pietraszek, Tadeusz},
  booktitle = {Recent Advances in Intrusion Detection (RAID)},
  year      = {2004},
}

@inproceedings{du2017deeplog,
  title     = {{DeepLog}: Anomaly Detection and Diagnosis from System Logs
               through Deep Learning},
  author    = {Du, Min and Li, Feifei and Zheng, Guineng and Srikumar, Vivek},
  booktitle = {Proceedings of the 2017 ACM SIGSAC Conference on Computer and
               Communications Security (CCS)},
  year      = {2017},
}

@inproceedings{vanede2022deepcase,
  title     = {{DeepCASE}: Semi-Supervised Contextual Analysis of Security
               Events},
  author    = {van Ede, Thijs and Aghakhani, Hojjat and Spahn, Noah and
               Bortolameotti, Riccardo and Cova, Marco and Continella, Andrea
               and van Steen, Maarten and Peter, Andreas and Kruegel,
               Christopher and Vigna, Giovanni},
  booktitle = {2022 IEEE Symposium on Security and Privacy (SP)},
  year      = {2022},
}

@inproceedings{milajerdi2019holmes,
  title     = {{HOLMES}: Real-Time {APT} Detection through Correlation of
               Suspicious Information Flows},
  author    = {Milajerdi, Sadegh M. and Gjomemo, Rigel and Eshete, Birhanu and
               Sekar, R. and Venkatakrishnan, V. N.},
  booktitle = {2019 IEEE Symposium on Security and Privacy (SP)},
  year      = {2019},
}

@misc{ali2023huntgpt,
  title         = {{HuntGPT}: Integrating Machine Learning-Based Anomaly
                   Detection and Explainable {AI} with Large Language Models},
  author        = {Ali, Tarek and Kostakos, Panos},
  year          = {2023},
  eprint        = {2309.16021},
  archivePrefix = {arXiv},
}

@misc{qi2023loggpt,
  title         = {{LogGPT}: Exploring {ChatGPT} for Log-Based Anomaly
                   Detection},
  author        = {Qi, Jiaxing and Huang, Shaohan and Luan, Zhongzhi and
                   Fung, Carol and Yang, Hailong and Qian, Depei},
  year          = {2023},
  eprint        = {2309.01189},
  archivePrefix = {arXiv},
}

@misc{albanese2025coteaming,
  title         = {Towards {AI}-Driven Human-Machine Co-Teaming for Adaptive and
                   Agile Cyber Security Operation Centers},
  author        = {Albanese, Massimiliano and Ou, Xinming and Lybarger, Kevin
                   and Lende, Daniel and Goldgof, Dmitry},
  year          = {2025},
  eprint        = {2505.06394},
  archivePrefix = {arXiv},
}

@misc{xiong2026sifting,
  title         = {Sifting the Noise: A Comparative Study of {LLM} Agents in
                   Vulnerability False Positive Filtering},
  author        = {Xiong, Yunpeng and Zhang, Ting},
  year          = {2026},
  eprint        = {2601.22952},
  archivePrefix = {arXiv},
}

@misc{jajodia2026siabench,
  title         = {Before You Hand Over the Wheel: Evaluating {LLMs} for
                   Security Incident Analysis},
  author        = {Jajodia, Sourov and Sultana, Madeena and Majumdar, Suryadipta
                   and Taylor, Adrian and Vandenberghe, Grant},
  year          = {2026},
  eprint        = {2603.06422},
  archivePrefix = {arXiv},
}

@misc{chen2023rcacopilot,
  title         = {Automatic Root Cause Analysis via Large Language Models for
                   Cloud Incidents},
  author        = {Chen, Yinfang and Xie, Huaibing and Ma, Minghua and Kang, Yu
                   and Gao, Xin and Shi, Liu and Cao, Yunjie and Gao, Xuedong
                   and Fan, Hao and Wen, Ming and Zeng, Jun and Ghosh, Supriyo
                   and Zhang, Xuchao and Zhang, Chaoyun and Lin, Qingwei and
                   Rajmohan, Saravan and Zhang, Dongmei and Xu, Tianyin},
  year          = {2023},
  eprint        = {2305.15778},
  archivePrefix = {arXiv},
}

@misc{zhang2023pacelm,
  title         = {{PACE-LM}: Prompting and Augmentation for Calibrated
                   Confidence Estimation with {GPT-4} in Cloud Incident Root
                   Cause Analysis},
  author        = {Zhang, Dylan and Zhang, Xuchao and Bansal, Chetan and
                   Las-Casas, Pedro and Fonseca, Rodrigo and Rajmohan, Saravan},
  year          = {2023},
  eprint        = {2309.05833},
  archivePrefix = {arXiv},
}

@misc{zhang2024llmcyber,
  title         = {When {LLMs} Meet Cybersecurity: A Systematic Literature
                   Review},
  author        = {Zhang, Jie and Bu, Haoyu and Wen, Hui and Liu, Yongji and
                   Fei, Haiqiang and Xi, Rongrong and Li, Lun and Yang, Yun and
                   Zhu, Hongsong and Meng, Dan},
  year          = {2024},
  eprint        = {2405.03644},
  archivePrefix = {arXiv},
}

@inproceedings{wei2022cot,
  title     = {Chain-of-Thought Prompting Elicits Reasoning in Large Language
               Models},
  author    = {Wei, Jason and Wang, Xuezhi and Schuurmans, Dale and
               Bosma, Maarten and Ichter, Brian and Xia, Fei and Chi, Ed H.
               and Le, Quoc V. and Zhou, Denny},
  booktitle = {Advances in Neural Information Processing Systems (NeurIPS)},
  year      = {2022},
}

@misc{suzgun2022bbh,
  title         = {Challenging {BIG-Bench} Tasks and Whether Chain-of-Thought
                   Can Solve Them},
  author        = {Suzgun, Mirac and Scales, Nathan and Sch{\"a}rli, Nathanael
                   and Gehrmann, Sebastian and Tay, Yi and Chung, Hyung Won and
                   Chowdhery, Aakanksha and Le, Quoc V. and Chi, Ed H. and
                   Zhou, Denny and Wei, Jason},
  year          = {2022},
  eprint        = {2210.09261},
  archivePrefix = {arXiv},
}

@misc{deepseekr1,
  title         = {{DeepSeek-R1}: Incentivizing Reasoning Capability in {LLMs}
                   via Reinforcement Learning},
  author        = {{DeepSeek-AI}},
  year          = {2025},
  eprint        = {2501.12948},
  archivePrefix = {arXiv},
}

@inproceedings{tian2023calibration,
  title     = {Just Ask for Calibration: Strategies for Eliciting Calibrated
               Confidence Scores from Language Models Fine-Tuned with Human
               Feedback},
  author    = {Tian, Katherine and Mitchell, Eric and Zhou, Allan and
               Sharma, Archit and Rafailov, Rafael and Yao, Huaxiu and
               Finn, Chelsea and Manning, Christopher D.},
  booktitle = {Proceedings of the 2023 Conference on Empirical Methods in
               Natural Language Processing (EMNLP)},
  year      = {2023},
}

@inproceedings{xiong2023confidence,
  title     = {Can {LLMs} Express Their Uncertainty? An Empirical Evaluation of
               Confidence Elicitation in {LLMs}},
  author    = {Xiong, Miao and Hu, Zhiyuan and Lu, Xinyang and Li, Yifei and
               Fu, Jie and He, Junxian and Hooi, Bryan},
  booktitle = {International Conference on Learning Representations (ICLR)},
  year      = {2024},
}

@inproceedings{zhou2023ape,
  title     = {Large Language Models Are Human-Level Prompt Engineers},
  author    = {Zhou, Yongchao and Muresanu, Andrei Ioan and Han, Ziwen and
               Paster, Keiran and Pitis, Silviu and Chan, Harris and Ba, Jimmy},
  booktitle = {International Conference on Learning Representations (ICLR)},
  year      = {2023},
}

@inproceedings{yang2024opro,
  title     = {Large Language Models as Optimizers},
  author    = {Yang, Chengrun and Wang, Xuezhi and Lu, Yifeng and Liu, Hanxiao
               and Le, Quoc V. and Zhou, Denny and Chen, Xinyun},
  booktitle = {International Conference on Learning Representations (ICLR)},
  year      = {2024},
}

@inproceedings{guo2024evoprompt,
  title     = {Connecting Large Language Models with Evolutionary Algorithms
               Yields Powerful Prompt Optimizers},
  author    = {Guo, Qingyan and Wang, Rui and Guo, Junliang and Li, Bei and
               Song, Kaitao and Tan, Xu and Liu, Guoqing and Bian, Jiang and
               Yang, Yujiu},
  booktitle = {International Conference on Learning Representations (ICLR)},
  year      = {2024},
}

@misc{khattab2023dspy,
  title         = {{DSPy}: Compiling Declarative Language Model Calls into
                   Self-Improving Pipelines},
  author        = {Khattab, Omar and Singhvi, Arnav and Maheshwari, Paridhi and
                   Zhang, Zhiyuan and Santhanam, Keshav and Vardhamanan, Sri and
                   Haq, Saiful and Sharma, Ashutosh and Joshi, Thomas T. and
                   Moazam, Hanna and Miller, Heather and Zaharia, Matei and
                   Potts, Christopher},
  year          = {2023},
  eprint        = {2310.03714},
  archivePrefix = {arXiv},
}

@misc{agrawal2025gepa,
  title         = {{GEPA}: Reflective Prompt Evolution Can Outperform
                   Reinforcement Learning},
  author        = {Agrawal, Lakshya A. and Tan, Shangyin and Soylu, Dilara and
                   Ziems, Noah and Khare, Rishi and Opsahl-Ong, Krista and
                   Singhvi, Arnav and Shandilya, Herumb and Ryan, Michael J. and
                   Jiang, Meng and Potts, Christopher and Sen, Koushik and
                   Dimakis, Alexandros G. and Stoica, Ion and Klein, Dan and
                   Zaharia, Matei and Khattab, Omar},
  year          = {2025},
  eprint        = {2507.19457},
  archivePrefix = {arXiv},
}

@misc{zelikman2024quietstar,
  title         = {{Quiet-STaR}: Language Models Can Teach Themselves to Think
                   Before Speaking},
  author        = {Zelikman, Eric and Harik, Georges and Shao, Yijia and
                   Jayasiri, Varuna and Haber, Nick and Goodman, Noah D.},
  year          = {2024},
  eprint        = {2403.09629},
  archivePrefix = {arXiv},
}

@misc{hosseini2024vstar,
  title         = {{V-STaR}: Training Verifiers for Self-Taught Reasoners},
  author        = {Hosseini, Arian and Yuan, Xingdi and Malkin, Nikolay and
                   Courville, Aaron and Sordoni, Alessandro and Agarwal, Rishabh},
  year          = {2024},
  eprint        = {2402.06457},
  archivePrefix = {arXiv},
}

@misc{gulcehre2023rest,
  title         = {Reinforced Self-Training ({ReST}) for Language Modeling},
  author        = {Gulcehre, Caglar and Paine, Tom Le and Srinivasan, Srivatsan
                   and Konyushkova, Ksenia and Weerts, Lotte and Sharma, Abhishek
                   and Siddhant, Aditya and Ahern, Alex and Wang, Miaosen and
                   Gu, Chenjie and Macherey, Wolfgang and Doucet, Arnaud and
                   Firat, Orhan and de Freitas, Nando},
  year          = {2023},
  eprint        = {2308.08998},
  archivePrefix = {arXiv},
}

@article{singh2024restem,
  title   = {Beyond Human Data: Scaling Self-Training for Problem-Solving with
             Language Models},
  author  = {Singh, Avi and Co-Reyes, John D. and Agarwal, Rishabh and others},
  journal = {Transactions on Machine Learning Research},
  year    = {2024},
}

@misc{yuan2023rft,
  title         = {Scaling Relationship on Learning Mathematical Reasoning with
                   Large Language Models},
  author        = {Yuan, Zheng and Yuan, Hongyi and Li, Chengpeng and
                   Dong, Guanting and Lu, Keming and Tan, Chuanqi and
                   Zhou, Chang and Zhou, Jingren},
  year          = {2023},
  eprint        = {2308.01825},
  archivePrefix = {arXiv},
}

@inproceedings{phan2023trice,
  title     = {Training Chain-of-Thought via Latent-Variable Inference},
  author    = {Phan, Du and Hoffman, Matthew D. and Dohan, David and
               Douglas, Sholto and Le, Tuan Anh and Parisi, Aaron and
               Sountsov, Pavel and Sutton, Charles and Vikram, Sharad and
               Saurous, Rif A.},
  booktitle = {Advances in Neural Information Processing Systems (NeurIPS)},
  year      = {2023},
}

@inproceedings{koh2025adastar,
  title     = {{AdaSTaR}: Adaptive Data Sampling for Training Self-Taught
               Reasoners},
  author    = {Koh, Woosung and Oh, Wonbeen and Jang, Jaein and Lee, MinHyung
               and Kim, Hyeongjin and Kim, Ah Yeon and Kim, Joonkee and
               Lee, Junghyun and Kim, Taehyeon and Yun, Se-Young},
  booktitle = {Advances in Neural Information Processing Systems (NeurIPS)},
  year      = {2025},
}

@inproceedings{ouyang2022instructgpt,
  title     = {Training Language Models to Follow Instructions with Human
               Feedback},
  author    = {Ouyang, Long and Wu, Jeff and Jiang, Xu and Almeida, Diogo and
               Wainwright, Carroll L. and Mishkin, Pamela and Zhang, Chong and
               Agarwal, Sandhini and Slama, Katarina and Ray, Alex and
               Schulman, John and Hilton, Jacob and Kelton, Fraser and
               Miller, Luke and Simens, Maddie and Askell, Amanda and
               Welinder, Peter and Christiano, Paul and Leike, Jan and
               Lowe, Ryan},
  booktitle = {Advances in Neural Information Processing Systems (NeurIPS)},
  year      = {2022},
}

@misc{schulman2017ppo,
  title         = {Proximal Policy Optimization Algorithms},
  author        = {Schulman, John and Wolski, Filip and Dhariwal, Prafulla and
                   Radford, Alec and Klimov, Oleg},
  year          = {2017},
  eprint        = {1707.06347},
  archivePrefix = {arXiv},
}

@misc{lambert2024tulu3,
  title         = {{T\"ulu 3}: Pushing Frontiers in Open Language Model
                   Post-Training},
  author        = {Lambert, Nathan and Morrison, Jacob and Pyatkin, Valentina
                   and Huang, Shengyi and Ivison, Hamish and Brahman, Faeze and
                   Miranda, Lester James V. and Liu, Alisa and Dziri, Nouha and
                   Lyu, Shane and Gu, Yuling and Malik, Saumya and Graf, Victoria
                   and Hwang, Jena D. and Yang, Jiangjiang and Le Bras, Ronan and
                   Tafjord, Oyvind and Wilhelm, Chris and Soldaini, Luca and
                   Smith, Noah A. and Wang, Yizhong and Dasigi, Pradeep and
                   Hajishirzi, Hannaneh},
  year          = {2024},
  eprint        = {2411.15124},
  archivePrefix = {arXiv},
}

@misc{kimiteam2025k15,
  title         = {{Kimi k1.5}: Scaling Reinforcement Learning with {LLMs}},
  author        = {{Kimi Team}},
  year          = {2025},
  eprint        = {2501.12599},
  archivePrefix = {arXiv},
}

@inproceedings{ahmadian2024rloo,
  title     = {Back to Basics: Revisiting {REINFORCE} Style Optimization for
               Learning from Human Feedback in {LLMs}},
  author    = {Ahmadian, Arash and Cremer, Chris and Gall{\'e}, Matthias and
               Fadaee, Marzieh and Kreutzer, Julia and Pietquin, Olivier and
               {\"U}st{\"u}n, Ahmet and Hooker, Sara},
  booktitle = {Annual Meeting of the Association for Computational
               Linguistics (ACL)},
  year      = {2024},
}

@incollection{platt1999,
  title     = {Probabilistic Outputs for Support Vector Machines and
               Comparisons to Regularized Likelihood Methods},
  author    = {Platt, John C.},
  booktitle = {Advances in Large Margin Classifiers},
  pages     = {61--74},
  publisher = {MIT Press},
  year      = {1999},
}

@inproceedings{guo2017calibration,
  title     = {On Calibration of Modern Neural Networks},
  author    = {Guo, Chuan and Pleiss, Geoff and Sun, Yu and Weinberger, Kilian Q.},
  booktitle = {International Conference on Machine Learning (ICML)},
  year      = {2017},
}

@misc{kadavath2022know,
  title         = {Language Models (Mostly) Know What They Know},
  author        = {Kadavath, Saurav and Conerly, Tom and Askell, Amanda and
                   Henighan, Tom and Drain, Dawn and Perez, Ethan and others},
  year          = {2022},
  eprint        = {2207.05221},
  archivePrefix = {arXiv},
}

@article{lin2022teaching,
  title   = {Teaching Models to Express Their Uncertainty in Words},
  author  = {Lin, Stephanie and Hilton, Jacob and Evans, Owain},
  journal = {Transactions on Machine Learning Research},
  year    = {2022},
}

@inproceedings{kuhn2023semantic,
  title     = {Semantic Uncertainty: Linguistic Invariances for Uncertainty
               Estimation in Natural Language Generation},
  author    = {Kuhn, Lorenz and Gal, Yarin and Farquhar, Sebastian},
  booktitle = {International Conference on Learning Representations (ICLR)},
  year      = {2023},
}

@misc{cobbe2021verifiers,
  title         = {Training Verifiers to Solve Math Word Problems},
  author        = {Cobbe, Karl and Kosaraju, Vineet and Bavarian, Mohammad and
                   Chen, Mark and Jun, Heewoo and Kaiser, Lukasz and
                   Plappert, Matthias and Tworek, Jerry and Hilton, Jacob and
                   Nakano, Reiichiro and Hesse, Christopher and Schulman, John},
  year          = {2021},
  eprint        = {2110.14168},
  archivePrefix = {arXiv},
}

@inproceedings{hu2022lora,
  title     = {{LoRA}: Low-Rank Adaptation of Large Language Models},
  author    = {Hu, Edward J. and Shen, Yelong and Wallis, Phillip and
               Allen-Zhu, Zeyuan and Li, Yuanzhi and Wang, Shean and
               Wang, Lu and Chen, Weizhu},
  booktitle = {International Conference on Learning Representations (ICLR)},
  year      = {2022},
}

@misc{nvidia2025nemotron3nano,
  title         = {{Nemotron 3 Nano}: Open, Efficient Mixture-of-Experts Hybrid
                   {Mamba}-Transformer Model for Agentic Reasoning},
  author        = {{NVIDIA}},
  year          = {2025},
  eprint        = {2512.20848},
  archivePrefix = {arXiv},
}

@misc{nvidia2026nemotron3super,
  title         = {{Nemotron 3 Super}: Open, Efficient Mixture-of-Experts Hybrid
                   {Mamba}-Transformer Model for Agentic Reasoning},
  author        = {{NVIDIA}},
  year          = {2026},
  eprint        = {2604.12374},
  archivePrefix = {arXiv},
}


\clearpage
\appendix

\section{Nemotron Models}

We use two open-weight Nemotron-3 models~\cite{nvidia2025nemotron3nano,nvidia2026nemotron3super}, both hybrid Mamba-2/attention mixture-of-experts (MoE) architectures with native \texttt{<think>} reasoning. \textbf{Nano} (30B total, 3B active) is both the classifier and the calibrator. \textbf{Super} (120B total, 12B active) is GEPA's reflection LM and the LLM judge for instruction quality and rationalization. Super is weaker at reasoning than frontier proprietary models, which we compensate for with a larger GEPA search budget.

\section{Prompt Optimization}

GEPA~\cite{agrawal2025gepa} ran with 100 candidate-equivalents, a reflection minibatch of 8, up to 15 merge invocations, and 16 threads ($\sim$166K metric evaluations). The task LM (Nano) generated at $T{=}1.0$ with a 1{,}024-token limit; the reflection LM (Super) used a 16{,}384-token limit. A custom proposer caps every candidate instruction at 1{,}024 tokens. The instruction-quality judge (Super) scores each candidate on five dimensions---multi-field reasoning, threshold rigidity, reasoning encouragement, field coverage, and generalization---each in $[0,1]$, and the optimization metric is $\mathrm{accuracy}\times\min(\mathrm{dimensions})$, so one weak dimension caps the score. The hand-written seed prompt below (a Charlotte~AI persona with per-field documentation) scored 62.0\% on validation:

\begin{lstlisting}[style=prompt]
You are Charlotte AI, an AI assistant with high knowledge in cybersecurity. You are specialized in helping the CrowdStrike customers with their detections within the Falcon platform.

You are given a detection document below which contains a list of fields related to a detection from CrowdStrike Falcon sensor. Your task is to choose the appropriate label (TRUE POSITIVE or FALSE POSITIVE) for the given detection. The documentation regarding the meaning of each field can be found below:
- `cmdline`: the command line used to create the target process.
- `filename`: the file name of the main executable for the target process.
- `filepath`: the full path to the main executable for the target process.
- `description`: a short description of the Indicator of Attack.
- `tactic` / `technique`: the MITRE ATT&CK tactic and technique.
- `severity`: the degree of severity given by the sensor.
- `local_prevalence` / `global_prevalence`: how widely the file was seen in the current customer's environment and across all customers.
- `parent_*` / `grandparent_*`: the corresponding fields for the parent and grandparent process.
- `overwatch_note`: a security assessment from CrowdStrike's elite threat-hunting team; highly valuable and should be prioritized.
- `pattern_disposition`: the remediation actions the sensor took (or should have taken) against a possible malicious process.
- `device_type`: Workstation, Server, or Domain Controller.
- `detection_description`: analytical reasoning explaining why the detection was generated.
- `precision`: a 0-100 percentage indicating how accurate the detection mechanism is overall.
- `threatpct`: a 0-100 percentage of detections confirmed as malicious threats.

Select the appropriate label based on the above fields.
- TRUE POSITIVE means it is malicious or performs an unwanted, dangerous action on the user's computer.
- FALSE POSITIVE means it is not malicious; it was flagged by mistake but is regular user usage.

Answer only with the label.
\end{lstlisting}

GEPA without the judge reached 87.5\% but converged on brittle numeric thresholds, whereas GEPA with the judge produced the 72.0\% prompt used downstream:

\begin{lstlisting}[style=prompt]
You are Charlotte AI, a cybersecurity assistant for CrowdStrike Falcon. Given a detection JSON containing fields such as cmdline, filename, filepath, description, tactic, technique, severity, local_prevalence, global_prevalence, parent_*, grandparent_*, overwatch_note, pattern_disposition, device_type, detection_description, precision, and threatpct, decide whether it is TRUE_POSITIVE or FALSE_POSITIVE. Evaluate all available evidence comparatively: weigh overwatch_note and detection_description for contextual insight, compare prevalence values (higher vs. lower) to assess rarity, examine parent and grandparent process lineage for legitimacy, consider the MITRE tactic and technique against the observed behavior, and factor in severity and threatpct as relative confidence indicators. No single field should deterministically decide the outcome; instead, collectively weigh the signals using relative language (e.g., more typical, less typical, higher likelihood, lower likelihood). Output only the label as TRUE_POSITIVE or FALSE_POSITIVE.
\end{lstlisting}

\section{Self-Training}

We finetune Nano with LoRA (rank 8, $\alpha{=}8$) for 25 AdaSTaR iterations (peak; 28 run), one epoch each, global batch 32, learning rate $5\times10^{-7}\to3\times10^{-6}$, sequences packed to 16{,}384 tokens, expert parallelism 8. Each iteration samples from a priority heap keyed by (last-sampled iteration, $K{=}10$ sliding-window win rate); the SFT-set cap grows as $s_i=\lfloor 2560\cdot1.2^{\,i-1}\rfloor$ and the per-iteration update budget is $\lfloor m\cdot\mathrm{acc}^2\rfloor$. Incorrect examples are \emph{rationalized}: we sample 3 hint-guided traces from the base model ($T{=}1.0$, two few-shot examples per class) and keep a trace only if a Super judge ($T{=}0.3$) scores it $\ge3$ on a 1--5 genuineness scale. The rationalization prompt and the judge prompt:

\begin{lstlisting}[style=prompt]
You are Charlotte AI, a CrowdStrike Falcon detection analyst. You will be given a detection with its correct classification. Your job is to write a detailed evidence-based analysis explaining why the detection is classified that way.

Your OUTPUT (after thinking) must be a thorough analysis that: (1) examines the cmdline and filename; (2) checks the parent/grandparent process chain; (3) notes pattern_disposition; (4) considers prevalence, precision, threatpct; (5) reaches a well-supported conclusion citing specific field values.

CRITICAL RULES:
- Do NOT reference or acknowledge that you were told the correct answer. Write as if you are determining the classification yourself from the evidence.
- Do NOT mention "the correct classification", "we are told", "given that this is", or any similar phrasing.
- Do NOT end your output with the classification label; output ONLY the evidence-based analysis. The label is appended separately.
\end{lstlisting}

\begin{lstlisting}[style=prompt]
You are evaluating a reasoning trace for a cybersecurity detection classification. The trace was produced by a model that was given the correct classification and asked to write an evidence-based analysis. Determine whether the trace reads as genuine independent analysis or reveals awareness of being told the answer.

Scoring:
1 = DISQUALIFIED. References being told the answer ("we are told", "the correct classification is"), OR opens with meta-reasoning ("We need to...", "Let me..."), OR is generic and cites no specific field values, OR is a bare label, OR is truncated.
2 = Superficial; mentions some fields but does not examine them in depth.
3 = Adequate; examines multiple fields with specific values. Minor issues tolerated.
4 = Strong; examines cmdline, parent context, disposition, and prevalence with specific values.
5 = Excellent; systematic evidence analysis citing specific field values and reaching a well-supported conclusion.

Output only the numeric score (1-5).
\end{lstlisting}

\section{Reinforcement Learning}

GRPO~\cite{deepseekmath2024} starts from the AdaSTaR checkpoint merged into the base weights, with a fresh LoRA adapter (rank 32, $\alpha{=}128$). Each step samples 48 prompts $\times$ 16 rollouts ($T{=}1.0$, 1{,}024-token limit); the binary reward is 1 if and only if the response is well-formatted \emph{and} correct. Advantages use a leave-one-out group baseline, and the clipped objective uses an asymmetric ratio clip $[0.2,0.28]$ with no KL penalty. Adam runs at a constant $3\times10^{-6}$ (10-step warmup); we validate every 50 steps on 512 samples and select step 8{,}000. Compute is 32 B200 GPUs in a non-colocated layout: Megatron-Core training (tensor parallel 2, expert parallel 4, data parallel 2) and vLLM generation (tensor parallel 4), with Ray orchestration and ZMQ weight synchronization.

\section{Confidence Calibration}

The calibrator is a separate Nano finetuned with LoRA (rank 32, $\alpha{=}32$, thinking disabled) to read a detection, the policy's reasoning trace, and its label, and emit a single token \texttt{correct} or \texttt{incorrect}. Writing $\ell_c, \ell_i$ for the two token log-probabilities, the probability of correctness is $p = e^{\ell_c}/(e^{\ell_c}+e^{\ell_i})$, and the reported confidence is $s = p$ if the verdict is \texttt{correct} and $s = 1-p$ otherwise. To form tiers, for each label we sort predictions by descending $s$ and take, as the threshold, the $s$ at the deepest rank whose cumulative precision still meets the target $\tau$; the high/medium/low tiers use the precision targets from the Experiments section. Thresholds are fit on validation and applied unchanged to test. Training data is 4 rollouts per detection ($T{=}1.0$), split 90/10 and balanced 50/50 by class, with label smoothing $\varepsilon{=}0.05$; we train 3 epochs (Adam, cosine $1\times10^{-5}\to1\times10^{-6}$, 100-step warmup), checkpoint every 50 steps, and find the strongest operating points at early checkpoints. The calibrator judge prompt:

\begin{lstlisting}[style=prompt]
You are a quality-assurance reviewer for cybersecurity incident classifications. You will be shown a security detection (raw incident data) followed by another analyst's reasoning and final verdict (TRUE_POSITIVE or FALSE_POSITIVE).

Your task: determine whether the analyst's classification is Correct or Incorrect. Respond with a single word: Correct or Incorrect.
\end{lstlisting}

\section{RLVR without Self-Training}

\begin{figure}[t]
\centering
\includegraphics[width=\columnwidth]{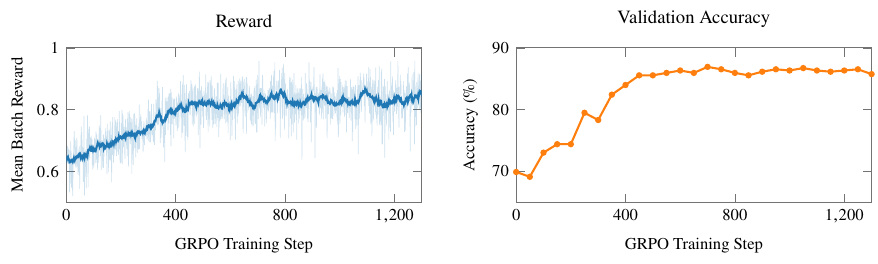}
\caption{RLVR without self-training: GRPO from the GEPA-prompted base model. Validation accuracy peaks at 86.9\%.}
\label{fig:rlvr}
\end{figure}

Figure~\ref{fig:rlvr} shows the \emph{+RLVR (no self-training)} ablation: GRPO applied directly to the GEPA-prompted base model, bypassing AdaSTaR. Validation accuracy rises from 69.9\% to a peak of 86.9\% and the mean reward climbs steadily, confirming that RLVR alone reaches strong policy accuracy on a well-prompted model.\footnote{This run matches the full pipeline's GRPO configuration on every major hyperparameter except the learning rate ($3\times10^{-6}$ vs.\ $1\times10^{-6}$). We modified learning rate to encourage quicker learning from a weaker (not self-trained) initial model.}

\end{document}